\documentclass{article}
\usepackage{ijcai24}

\usepackage{times}
\usepackage{soul}
\usepackage{url}
\usepackage[hidelinks]{hyperref}
\usepackage[utf8]{inputenc}
\usepackage[small]{caption}
\usepackage{graphicx}
\usepackage{amsmath}
\usepackage{amsthm}
\usepackage{booktabs}
\usepackage{algorithm}
\usepackage{algorithmic}
\usepackage[switch]{lineno}

\usepackage{color}
\usepackage[cmyk]{xcolor} 

\title{\textit{We Choose to Go to Space}: Agent-driven Human and Multi-Robot Collaboration in Microgravity}

\author{
    Author Name \footnote{asdas}
    \affiliations
    Affiliation
    \emails
    email@example.com
}

\author{
Miao Xin$^{1}$\footnote{Corresponding author}
\and
Zhongrui You$^{2\dag}$
\and
Zihan Zhang$^{3\dag}$
\and
Taoran Jiang$^{4\dag}$
\and
Tingjia Xu$^{5\dag}$
\and
Haotian Liang$^{4\dag}$
\and
Guojing Ge$^1$
\and
Yuchen ji$^{6\dag}$
\and
Shentong Mo$^7$
\and
Jian Cheng$^1$\\
\affiliations
$^1$Institute of Automation Chinese Academy of Sciences \enspace  
$^2$Beijing University of Posts and Telecommunications \enspace
$^3$TongJi University  \enspace
$^4$Beihang University \enspace
$^5$Beijing Normal University \enspace
$^6$Nanjing University of Aeronautics and Astronautics \enspace
$^7$Carnegie Mellon University 
} 

\begin{document}

\twocolumn[{%
\renewcommand\twocolumn[1][]{#1}%
\maketitle
\begin{center}
\centering
\includegraphics[width=\textwidth]{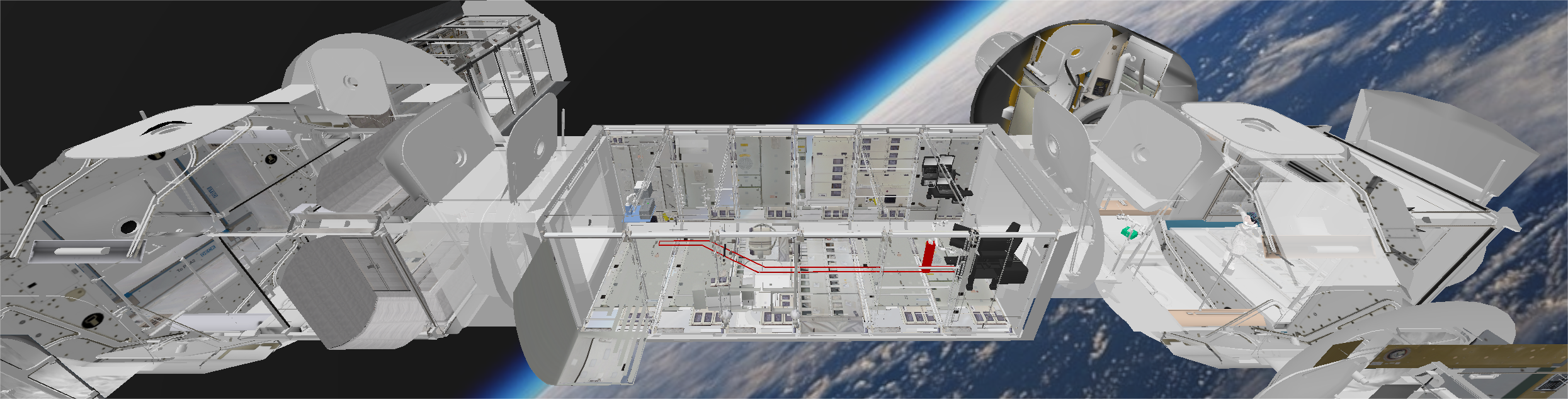}
\captionof{figure}{\textbf{SpaceAgents-1} simulates the human and multi-robot collaboration in microgravity under the control of AI Agents. In this figure, an astronaut is transporting cargoes across multiple modules with a free-flying robot, a rail-type robot and a dexterous robot working together. }  
\label{fig:banner} 
\end{center}%
}]

\begin{abstract}
We present \textbf{SpaceAgents-1}, a system for learning human and multi-robot collaboration (HMRC) strategies under microgravity conditions. 
Future space exploration requires humans to work together with robots. 
However, acquiring proficient robot skills and adept collaboration under microgravity conditions poses significant challenges within ground laboratories. 
To address this issue, we develop a microgravity simulation environment and present three typical configurations of intra-cabin robots. 
We propose a hierarchical heterogeneous multi-agent collaboration architecture: guided by foundation models, a Decision-Making Agent serves as a task planner for human-robot collaboration, while individual Skill-Expert Agents manage the embodied control of robots. 
This mechanism empowers the SpaceAgents-1 system to execute a range of intricate long-horizon HMRC tasks.  
\end{abstract}

\section{Introduction}
\label{sect:introduction}

\footnotetext{$^{\star}$Corresponding author (e-mail: miao.xin@ia.ac.cn). $\dag$Work done while they were Research Interns with CASIA.  } 
We believe that human beings will eventually go into further space and to the outer planets. 
In the current maintenance of space stations~\cite{Li2022ChinasSR} and future space exploration~\cite{Ackerman2022RoboticsMT}, a great deal of missions need to be carried out by both humans and robots. 
This makes embodied collaborations between machine intelligence and human intelligence essential.  
However, due to the particularity of microgravity, counterintuitive phenomena are difficult to obtain via experiments on Earth. 
Therefore, highly realistic simulation of microgravity and learning of HMRC is crucial.

In this work, we demonstrate the \textit{SpaceAgent-1} system, designed to learn robot skills in microgravity, and explore the HMRC to accomplish various tasks inside the cabin. 
The system is driven by AI Agents~\cite{durante2024agent}. 
We designed a hierarchical multi-agent architecture to achieve collaborative control and embodied execution. 
A foundation model~\cite{openai2023gpt4} provides the agents with multiple abilities such as planning, memory, action, and critique, allowing them to exhibit the potential to align with human skills. 
We also develop a platform, \textit{SpaceSim}, for microgravity physics simulation and provide space robot simulators.

In summary, we explore a potential technical route that coordinates HMRC~\cite{Lippi2022ATA} in microgravity employing AI Agents and provide a simulation platform for verification, presenting a new endeavor for future research.

\section{\textit{SpaceSim}: Microgravity Simulation}
\label{sect:system}

The SpaceSim platform consists of three parts: \textit{i.} microgravity simulator, \textit{ii.} space robots, and \textit{iii.} real2sim human-robot interaction (HRI).

\textbf{Microgravity simulation}. 
The 3D environment provides highly realistic physical simulations of microgravity and various phenomena it causes. 
The simulation engine of SpaceSim is built on SAPIEN~\cite{Xiang2020SAPIENAS}, a part-based and physics-rich simulation environment that allows us to train robot skills based on reinforcement learning (RL) through the Gym interface.

\begin{figure}[ht]
	\centering 
	\includegraphics[width=\linewidth]{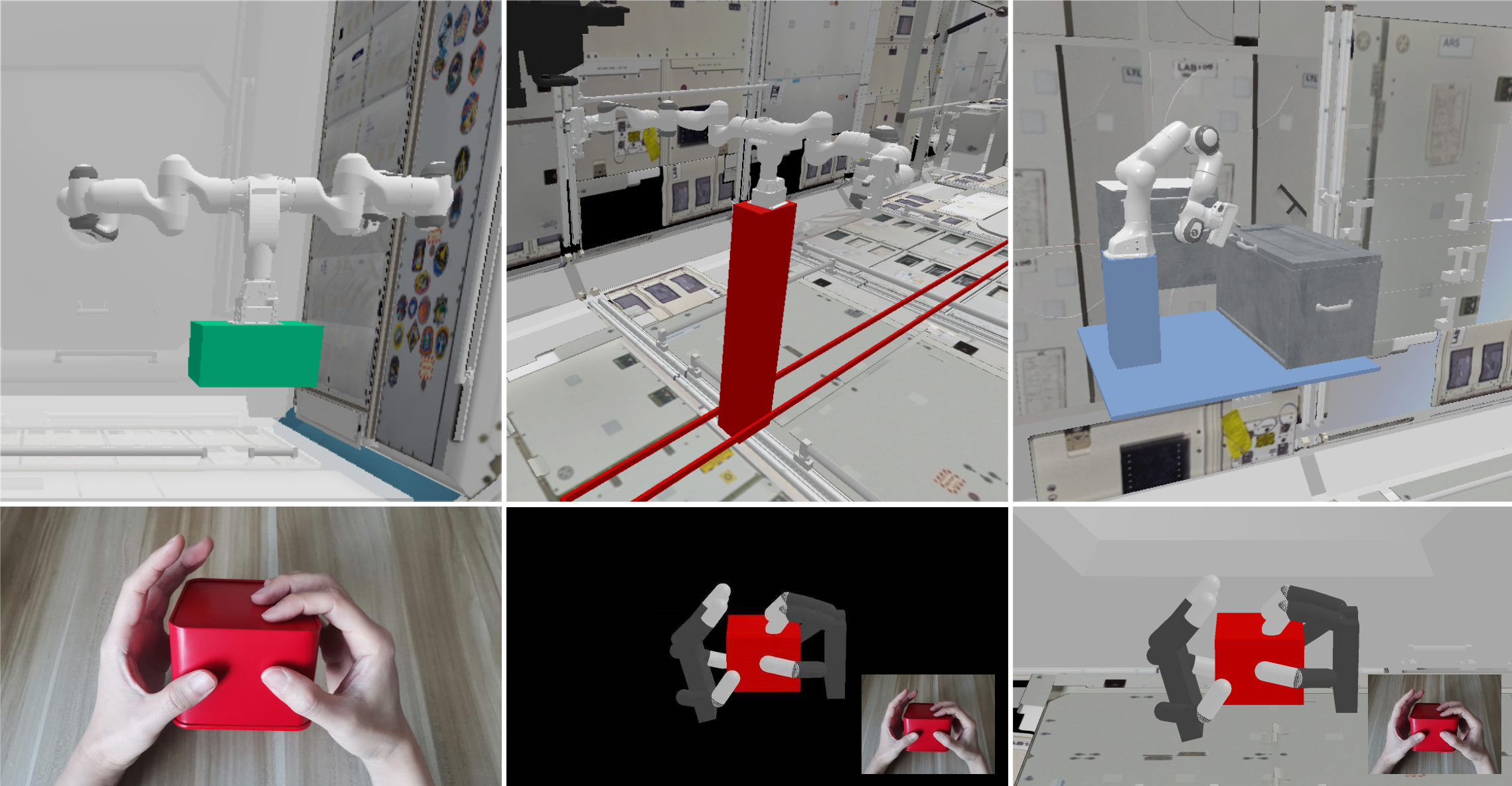} 
	\caption{Robots in SpaceSim (up) and HRI simulation (bottom).}  
	\label{fig:robot_type}  
\end{figure}

\textbf{Intra-cabin space robots}. 
SpaceSim provides three types of robots (Figure~\ref{fig:robot_type}).  
(1) \textbf{\textcolor{green}{Free-flying robot (F)}} can fly aerodynamically inside the space station, similar to \textit{Astrobee}~\cite{Bualat2018AstrobeeAN} robot but with limited cross-cabin dragging capability. 
(2) \textbf{\textcolor{red}{Rail-type robot (R)}} can move along the sliding rail and grasp objects with the robot arms, which is suitable for stable handling tasks within one module. 
(3) \textbf{\textcolor{blue}{Dexterous robot (D)}} can perform delicate operational tasks, such as \textit{picking} objects, or \textit{opening}  boxes.

\textbf{Real2sim HRI interface}. 
With the hand-object interaction algorithm~\cite{Zhou2020MonocularRH}, \textbf{\textcolor{orange}{Human (H)}} collaborators map real-world hand manipulations to the SpaceSim environment through visual teleoperation. 
Compared to using specialized equipment~\cite{Glauser2019InteractiveHP}, computer vision-based hand-object motion capture has lower costs, enabling large-scale data collection.

\section{\textit{SpaceAgents}: Hybrid and Embodied Multi-agent Collaboration }
\label{sect:agent}

\subsection{Hierarchical Architecture}

HMRC typically performs in human-supervisor and robot-executor paradigm, or equal partnership paradigm~\cite{Xi2023TheRA}. 
In contrast to both of these, we adopt a hierarchical mode.
The \textbf{Decision-Making Agent (DMA)} controls the collaboration between various \textbf{Skill-Expert Agents (SEAs)} and humans, while each embodied SEA controls its corresponding robot. 
DMA focuses on global collaborative planning, but it is not embodied. 
In contrast, SEAs can accomplish embodied execution, but are more focused on executing specific skills. 
The advantage of this design lies in the decoupling of policies. 
When a certain SEA needs to acquire new skills, the system does not need to be updated as a whole.

\subsection{Working Flow}
\label{sect:working_flow}

\begin{figure}[ht]
	\centering 
	\includegraphics[width= \linewidth]{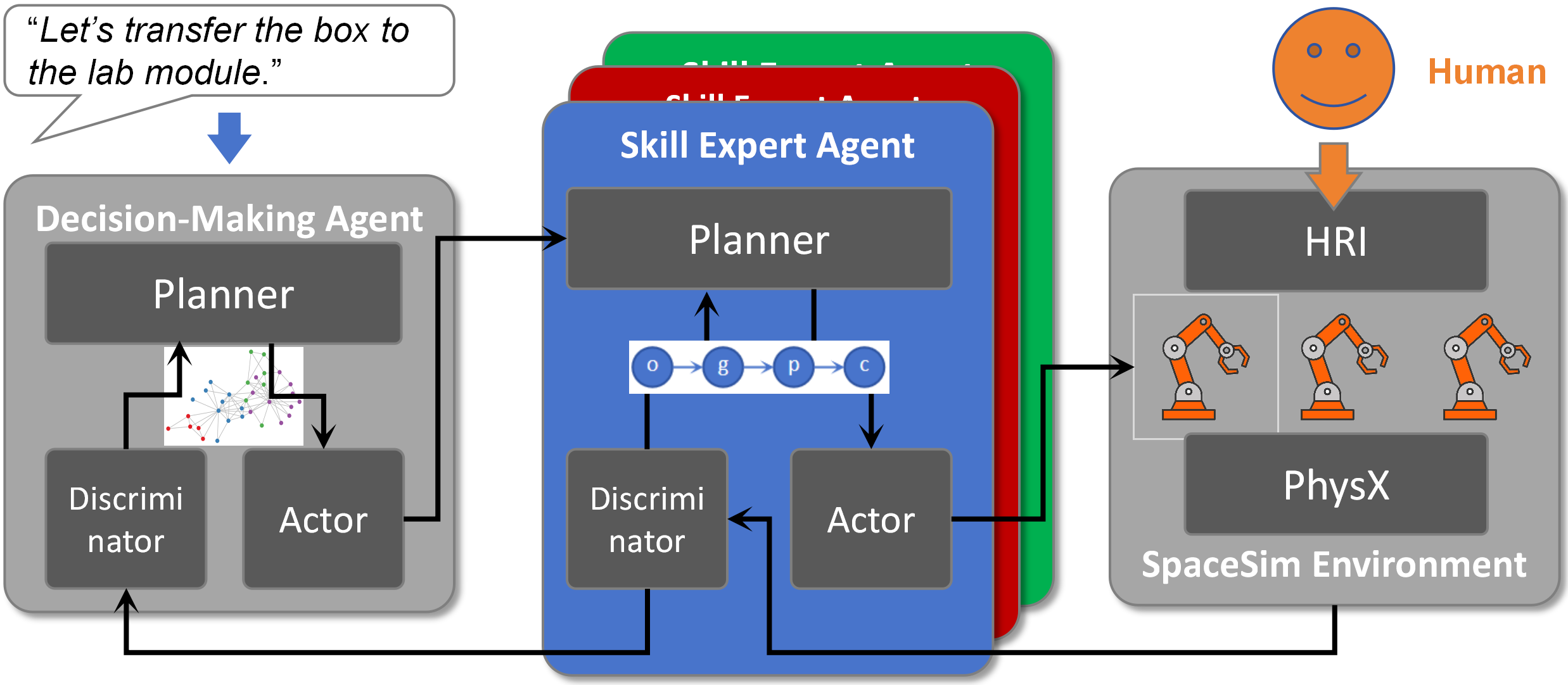} 
	\caption{System working flow. }  
	\label{fig:working_flow} 
\end{figure}

The Agent consists of a \textit{Planner}, an \textit{Actor} and a \textit{Discriminator}. 
Figure~\ref{fig:working_flow} sketches the working flow of the system.

\begin{figure}[ht]
	\centering 
	\includegraphics[width=\linewidth]{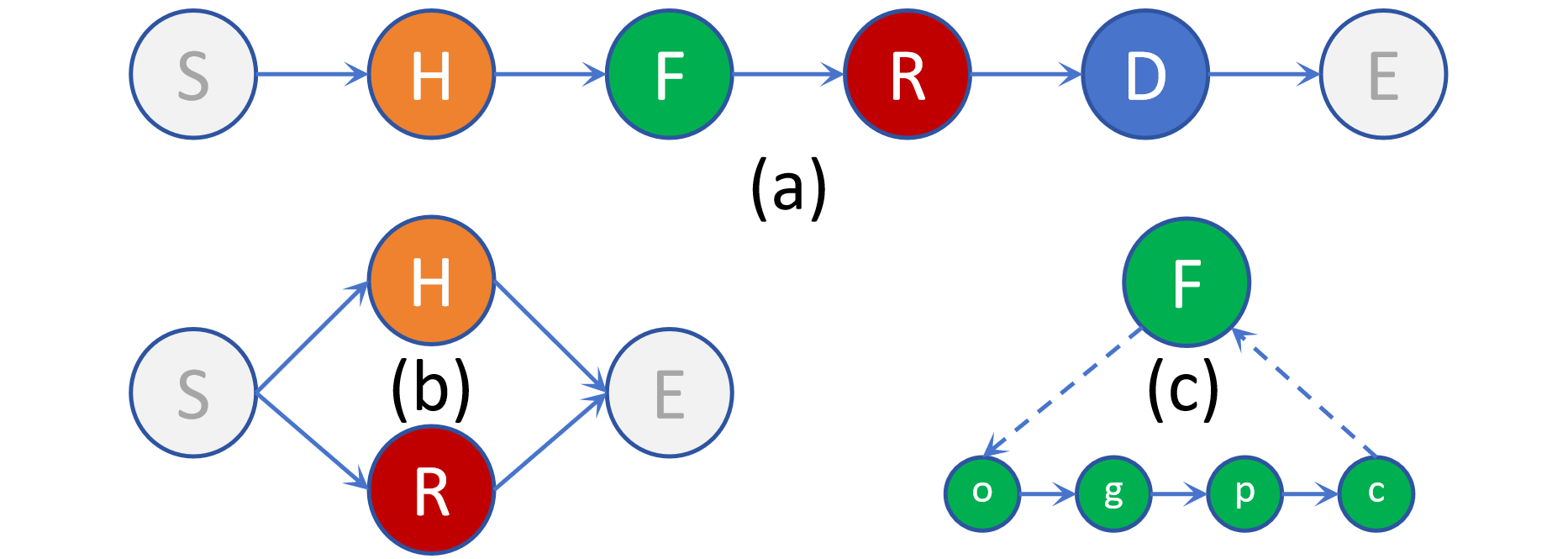} 
	\caption{Serial (a) and parallel (b) subtask collaboration graphs. (c) A subtask is decomposed into a basic skill chain. }  
	\label{fig:graph} 
\end{figure}

\textbf{Collaborative Planning.}
Given a long-horizon task described in natural language, the DMA Planner first decomposes it into multiple subtasks and allocates each subtask to either a SEA (preferentially) or a human executor based on estimated affordance.  
To express the collaborative relationships among these subtasks, the DMA constructs a directed \textit{collaboration graph} (CoG, Figure~\ref{fig:graph}) within its working memory. 
The capability for collaborative task planning originates from the foundation model~\cite{openai2023gpt4}, which directly outputs the collaboration graph in JSON format guided by prompt engineering~\cite{Mo2023TreeOU}. 
CoGs can represent both serial and parallel collaborations. 
Serial collaboration refers to a sequential order between task nodes, where a task node must wait until the preceding one is completed.
Conversely, parallel nodes allow multiple subtasks to start simultaneously, with the next subtask initiating only after all preceding subtasks have concluded.

Subtasks are described in text instructions.
For one subtask instruction, a Planner in the designated SEA  further decomposes it into a \textit{skill chain}~\cite{Mishra2023GenerativeSC} consisting of basic skills (\textit{e.g.}, ``\textit{(g)rasp}'', ``\textit{(o)pen}'', ``\textit{(p)ick}'', etc.). 
A basic skill is an RL policy learned via Proximal Policy Optimization (PPO)~\cite{PPO}, which resides in the long-term memory (skill library) of the SEA.

\textbf{Task Execution.} 
Actors transform instructions into actions. 
DMA Actor maintains communication between SEAs and ensures that the security boundaries of each SEA are not violated during execution. 
For each skill instruction in a skill chain, the Actor in a SEA invokes the corresponding RL policy to output the skill action vector directly to the robot. 
We utilize reward engineering to facilitate skill smoothing. 
The observation and training objectives of RL include not only the position of the termination point but also the instantaneous state at the termination of the skill, such as the velocity, acceleration, and 6-DoF pose of the manipulated object. 
This ensures a smooth connection between the end of one skill and the beginning of the next.

\textbf{Result Evaluation.}  
Given the task instruction, SEA Discriminator generates a structured state description of current observations with the help of a vision-language model~\cite{liu2023llava} and identifies whether the current state matches the goal of the task. 
The Discriminator of DMA collects messages from each SEA Discriminator in a multi-process manner and provides the DMA Planner with judgments on whether the tasks of each collaborator are accomplished.

With this closed workflow, planning, execution, and discrimination form an agent-driven HMRC loop, navigating collaborators to traverse the collaboration graph sequentially and ultimately completing the task.

\section{Long-horizon HMRC Manipulation}
\label{sect:demo}

In our \textbf{demo video}$^1$\footnote{$^1$Youtube Video  (\url{https://youtu.be/GLSjUUtp32k})}, we demonstrate certain interesting phenomena of weightlessness and basic skills of robots in the simulation environment.  
Specifically, we demonstrate two representative long-horizon HMRC cases.

\subsection{Case 1: Floating Objects Rearrangement}
\label{sect:demo_1}

\begin{figure}[ht]
	\centering 
	\includegraphics[width=\linewidth]{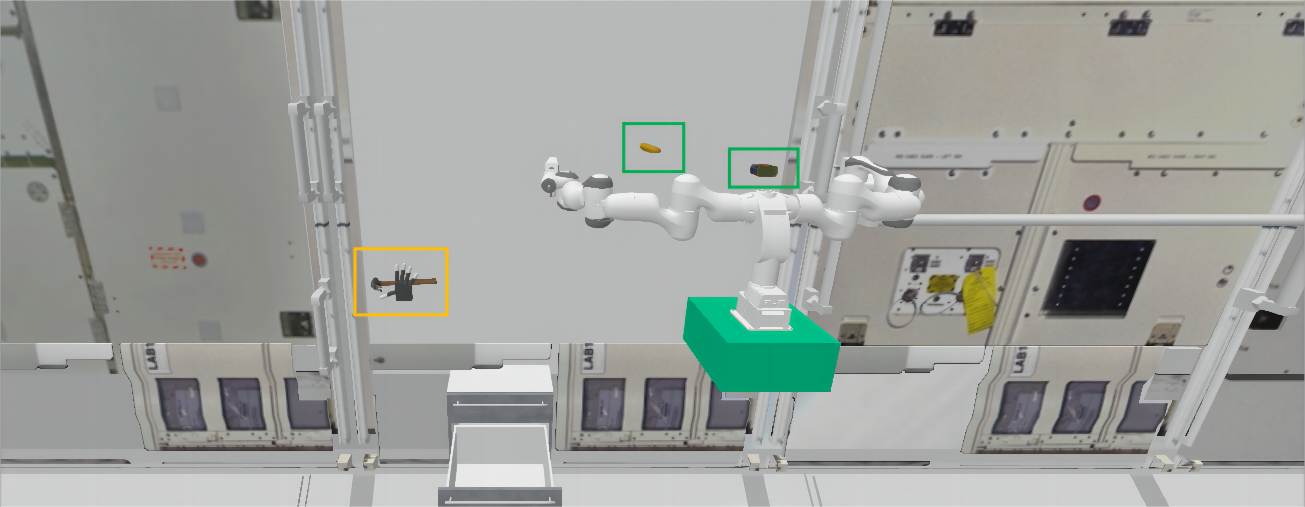} 
	\caption{Floating objects rearrangement. }  
	\label{fig:demo_1} 
\end{figure}

In this demo (Figure~\ref{fig:demo_1}), humans and robots work together to collect objects floating in the space station.  
To simplify, Planners use the object state information provided by the simulator, such as spatial position and size. 
DMA assigns subtasks to humans or robots based on the distance between the floating objects and the executors, thereby ensuring that there is no mutual interference between them.

\subsection{Case 2: Relay Object Transport}
\label{sect:demo_2}

In long-horizon planning, Agents not only seek to minimize human workload as much as possible but also comprehensively consider the skill availability of different executors and environmental constraints. 
The prompts provide affordance information about the capabilities and limitations of different robots. 
For example, the rail-type robot can stably grasp objects but can only move along rails within a single module. 
The free-flying robot can pass through cabin connections but struggles with stably dragging objects over long distances. 
The dexterous robot can perform delicate operations but is limited to objects fixed on the tabletop. 
In this case (Figure~\ref{fig:banner}), humans collaborate with multiple robots under the control of SpaceAgents-1 to complete the task of transporting cargo between multiple modules, effectively reducing human labor.

\subsection{Results}

In Table~\ref{tab:basic_skill}, we compare the success rates of the same dexterou robot completing identical tasks under microgravity or gravity conditions~\cite{gu2023maniskill2}. 
We observe that it is more difficult for robots to complete tasks in microgravity. 
It is almost impossible for robots to successfully utilize existing skill strategies directly in microgravity, which reflects the importance of learning in a simulation environment. 
The success rates of grasping skills for Free-flying robots and Rail-type robots reach $0.84$.

\begin{table}[ht]
\small 
\setlength\tabcolsep{3pt}
    \centering
    \begin{tabular}{lccccccc}
        \toprule
        \textbf{Robots}  & \textbf{P}   & \textbf{T}     & \textbf{CF} & \textbf{OC} & \textbf{CC} & \textbf{OB} & \textbf{PB}     \\
        \midrule
        Dexterous (0g)   & 0.88 & 0.93 & 0.72 & 0.79 & 0.96 & 0.93 & 0.78     \\
        Dexterous (1g)   & 0.91 & 0.98 & - & 0.92 & 0.95 & 0.93 & 0.82     \\
        \bottomrule
    \end{tabular}
    \caption{Success rates of basic skills (\textbf{P}ickCube, \textbf{T}hrowCube, \textbf{C}atch\textbf{F}loatingCube, \textbf{O}pen\textbf{C}abinetDoor, \textbf{C}lose\textbf{C}abinetDoor, \textbf{O}pen\textbf{B}ox, \textbf{P}ickfrom\textbf{B}ox) in microgravity (0g) or earth  gravity (1g). ``-'' represents the robot does not have this skill.}
    \label{tab:basic_skill}
\end{table}

We also compare the performance of SpaceAgents-1 with that of human experts (Table~\ref{tab:agent_vs_human}). 
With teleoperation, individuals act as the controllers of the robot instead of the agent. 
Across two collaborative tasks, SpaceAgents-1 demonstrates human-comparable performance in long-horizon task decomposition, attributable to the capabilities of the foundation model. 
However, regarding skill execution, humans encounter challenges in stably controlling robots in microgravity without prior training, contrasting starkly with terrestrial conditions. 
Nonetheless, a considerable number of failures are observed in the experiment, highlighting the irreplaceability of neither robots nor humans.

\begin{table}[ht]
\small 
    \centering
    \begin{tabular}{lcc}
        \toprule
        \textbf{Tasks}                  & Experts   & SpaceAgents-1   \\
        \midrule
        Floating Objects Rearrangement  & 0.91  & 0.87    \\
        Relay Object Transport          & 0.46  & 0.39    \\
        \bottomrule
    \end{tabular}
    \caption{Success rates on long-horizon tasks.}
    \label{tab:agent_vs_human}
\end{table}

\section{Conclusion}

This paper presents SpaceAgents-1, an HMRC system in microgravity. 
We design a hierarchical multi-agent collaboration architecture, which is driven by the foundation model to complete various tasks such as collaborative planning, embodied execution and effect evaluation. 
The space physics simulation environment we built allows the robots to learn operational skills and collaborative processes in microgravity at a very low cost.
We will open-source this system in future work, hoping that this effort can inspire more researchers.

\bibliographystyle{named}
\bibliography{ijcai24}

\end{document}